\newcommand{\CLASSINPUTbaselinestretch}{1}
\documentclass[conference]{IEEEtran}
\usepackage{etoolbox}
\makeatletter
\patchcmd{\maketitle}{\@copyrightspace}{}{}{}

\makeatother
\usepackage{xspace}
\usepackage{siunitx}
\usepackage[]{graphicx}
\usepackage{stfloats}  
\usepackage{cite}  
\usepackage{psfrag}  
\usepackage{subfigure}
\usepackage{wrapfig}
\usepackage{epsfig}
\usepackage{url}
\usepackage{amsmath, amssymb}
\usepackage{array}
\usepackage{multicol}
\usepackage{algorithm}
\usepackage{algorithmic}
\usepackage{mathrsfs}
\usepackage{multirow}
\usepackage[normalem]{ulem}
\usepackage{verbatim}
\usepackage{fixltx2e}
\usepackage[utf8x]{inputenc} 
\usepackage{ucs} 
\usepackage{amsfonts} 
\usepackage{makeidx} 
\usepackage[dvipsnames,usenames]{color} 
\usepackage{array} 
\usepackage{colortbl} 
\usepackage{booktabs} 
\usepackage[justification=centering]{caption}

\usepackage{pifont}
\newcommand{\cmark}{\ding{51}}%
\newcommand{\xmark}{\ding{55}}%
\newcommand{\name}{TxSim}%
\definecolor{tableheading}{rgb}{0.9,0.9,0.9}
\definecolor{softblue}{rgb}{0.8,0.8,1} 
\arrayrulecolor{Black} 
\usepackage{tabu}
\usepackage{floatflt}
\usepackage{amsmath, amssymb}
\usepackage{tikz}
\usepackage{multicol}
\usepackage{amsmath,amssymb}
\usepackage{tabularx}
\usepackage{soul}

\begin{document}

\title{
  \name: Modeling Training of Deep Neural Networks on Resistive Crossbar Systems\vspace{-0.25in}}
\author{\IEEEauthorblockN{Sourjya Roy\textsuperscript{*}, Shrihari Sridharan\textsuperscript{*}, Shubham Jain\textsuperscript{\textdagger} and Anand Raghunathan}
School of Electrical and Computer Engineering, Purdue University, West Lafayette, IN 47906 USA\\ }
\maketitle
\begingroup\renewcommand\thefootnote{\textsection}
\footnotetext{\textsuperscript{*}Equal contribution listed in alphabetical order}
\footnotetext{{\textdagger Shubham Jain is currently a research staff member at IBM T.J. Watson Research
Center, Yorktown Heights, NY (shubham.jain35@ibm.com)}}
\endgroup


\sloppy

\maketitle

\begin{abstract}
Deep Neural Networks (DNNs) have gained tremendous popularity in recent years due to their ability to achieve superhuman accuracy in a wide variety of machine learning tasks. However, the compute and memory requirements of DNNs have grown rapidly, creating a need for energy-efficient hardware. Resistive crossbars have attracted significant interest in the design of the next generation of DNN accelerators due to their ability to natively execute massively parallel vector-matrix multiplications within dense memory arrays. However, crossbar-based computations face a major challenge due to device and circuit-level non-idealities, which manifest as errors in the vector-matrix multiplications and eventually degrade DNN accuracy. To address this challenge, there is a need for tools that can model the functional impact of non-idealities on DNN  training and inference. Existing efforts towards this goal are either limited to inference, or are too slow to be used for large-scale DNN training. 
\\ \indent We propose \name, a fast and customizable modeling framework to functionally evaluate DNN training on crossbar-based hardware considering the impact of non-idealities. The key features of~\name~that differentiate it from prior efforts are: (i) It comprehensively models non-idealities during all training operations (forward propagation, backward propagation, and weight update) and (ii) it achieves computational efficiency by mapping crossbar evaluations to well-optimized BLAS routines and incorporates speedup techniques to further reduce simulation time with minimal impact on accuracy. \name~achieves significant improvement in simulation speed over prior works, and thereby makes it feasible to evaluate training of large-scale DNNs on crossbars. Our experiments using \name~reveal that the accuracy degradation in DNN training due to non-idealities can be substantial (3\%-36.4\%) for large-scale DNNs and datasets, underscoring the need for further research in mitigation techniques. We also analyze the impact of various device and circuit-level parameters and the associated non-idealities to provide key insights that can guide the design of crossbar-based DNN training accelerators. 
\end{abstract}

\section{Introduction}
\label{sec:introduction}
{\noindent} Deep Neural Networks (DNNs) have greatly advanced the state-of-the-art in a wide variety of machine learning tasks~\cite{fortune-dnns,speech}. However, these benefits come at the cost of extremely high computation and storage requirements. GPUs and digital CMOS-based accelerators~\cite{tpu,brainwave} have enabled faster and more energy-efficient realization of DNNs. However, the continuing growth in network complexities and volumes of data processed have led to the quest for further improvements in hardware. For example, training state-of-the-art DNNs requires exa-ops of compute and can take days to weeks on a GPU~\cite{scaledeep}, while Neural Architecture Search (NAS)~\cite{nas} further increases computation requirements to zetta-ops.

Resistive Crossbars have emerged as promising building blocks for future DNN accelerators. They are designed using emerging non-volatile memory technologies such as PCM~\cite{pcm} and ReRAM\cite{reram} that can enable high-density memory arrays, while also realizing massive parallel vector-matrix multiplications (the dominant compute kernel of DNNs) within these arrays. Thus, crossbar-based architectures promise to overcome the data transfer and memory capacity bottlenecks that are present in current DNN hardware platforms. Many efforts have explored the design of crossbar-based accelerators~\cite{neuromorphicHardware_Survey,ibmall}. We specifically focus on crossbar-based architectures for DNN training~\cite{pipelayer,crossim,tayfunMLP}, which have attracted increasing interest in recent years.

Crossbar-based systems face a major challenge due to numerous device and circuit-level non-idealities,~\emph{viz.}, driver and sensing resistances, analog-to-digital converter (ADC) and digital-to-analog converter (DAC) non-linearity, interconnect resistances, process variations, imperfect write  operations, and sneak paths\cite{neurosim,pytorx,rxnn,crossim}. Unless addressed, these non-idealities can significantly degrade DNN accuracy, threatening the viability of crossbar-based hardware~\cite{ibmall}. To quantitatively evaluate and address this challenge, there is a need for tools that can model the impact of all non-idealities on each step of DNN training (forward propagation, backward propagation, and weight update). DNN training on native hardware (e.g., GPUs) is already a bottleneck, and software simulation of crossbar-based DNN training  systems with detailed modeling of non-idealities will be much slower. Therefore, it is extremely important that the modeling tool maintains high simulation speed (same order-of-magnitude as DNN training on native hardware). Further, the tool should also be customizable and support a wide variety of device and circuit parameters and DNN topologies. 

In this work, we propose TxSim, a tool to functionally evaluate DNN training on crossbar-based systems, which meets the aforementioned requirements. TxSim utilizes a three-stage crossbar model to capture the impact of non-idealities during forward and backward propagation operations with good fidelity and simulation speed. The first stage consists of a non-linear conversion of digital inputs to voltages considering DAC non-idealities. The second stage models the non-idealities within the core crossbar array (interconnect parasitics, sneak paths and process variations) as a series of linear-algebraic transformations wherein ideal conductance matrices are converted to non-ideal conductance matrices. The non-ideal conductance matrices are in turn used to map input voltages into output currents. The final stage consists of the non-linear transformation of the currents back to digital outputs considering ADC non-idealities. Such an approach to modeling allows us to seamlessly utilize highly-optimized BLAS routines present in standard ML frameworks (\emph{e.g.}, PyTorch, Tensorflow and Caffe) to perform crossbar evaluation. TxSim also models the non-idealities involved in weight update operations (stochastic noise and update non-linearity). Finally, TxSim proposes speedup techniques that further reduce simulation time without impacting modeling fidelity.

Prior efforts on functional modeling of crossbar-based DNN hardware can be broadly classified into efforts that model inference~\cite{neurosim,pytorx,rxnn} and efforts that model training~\cite{neurosim,crossim}. Inference models are not sufficient for evaluating DNN training, since training includes additional backward propagation and weight update operations. 
As elaborated in Section~\ref{sec:relatedWork}, TxSim's modeling approach and speedup techniques make it significantly faster than prior efforts to model DNN training on crossbars~\cite{neurosim,crossim}. It achieves this while also being more comprehensive in the non-idealities modeled ({\em e.g.}, wiring parasitics), and being customizable to different DNN topologies and circuit and device parameters. In summary, our key contributions are:

\begin{itemize}
\item We propose TxSim, a scalable and customizable modeling framework to functionally evaluate DNN training on crossbar-based system. TxSim models a more comprehensive set of non-idealities and is significantly faster than prior training frameworks. 
\item We introduce speedup techniques utilizing approximate but high fidelity models to further improve simulation speeds.
\item We analyze the impact of various device and circuit-level parameters and the associated non-idealities on DNN training and provide key insights to guide cross-layer optimizations for crossbar-based DNN training accelerators. 
\end{itemize}

The rest of the paper is organized as follows. Section II discusses the previous works that model crossbar based architectures. Section III provides a brief overview of DNN training and background on resistive crossbar based systems. Section IV presents the TxSim modeling tool and details its various components. Section V discusses the evaluation methodology. Section VI quantifies the application level accuracy degradation of DNN training on resistive crossbars and provides sensitivity analysis to various circuit and device level parameters and Section VII concludes the paper.

\vspace*{-0pt}

\vspace*{2pt}
\section{Related Work }
\label{sec:relatedWork}
In this section, we discuss prior efforts to modeling inference and training on crossbar-based systems, as well as training algorithms/methodologies for such systems.

\noindent \textbf{Inference} \textbf{modeling. } ~\cite{pytorx},~\cite{rxnn},~\cite{mnsim2},~\cite{noiseinj} are modeling tools that consider the impact of non-idealities in crossbar-based inference. Other works~\cite{Chakraborty_2018,cdnn} propose methods to compensate for accuracy degradation. However, these tools are not directly applicable to crossbar-based training, which involves modeling the non-idealities in the backward propagation and weight update phases as well.  

\noindent\textbf{Area, performance and energy modeling.} Various efforts that propose crossbar-based training architectures~\cite{pipelayer,crossim,tayfunMLP} also develop performance and energy models to evaluate them. MNSIM \cite{mnsim} is a tool for early design space exploration of such architectures. The major focus of these works have been on area, speed and energy while ignoring or assuming very primitive error models for non-idealities.

\noindent\textbf{Modeling crossbar-based training.} Two noteworthy efforts that model crossbar-based training are CrossSim~\cite{crossim} and  NeuroSim~\cite{neurosim}. Table~\ref{tab:lim_control} compares our work with these efforts along two important dimensions -- the fidelity in modeling non-idealities and the simulation time. CrossSim considers only non-idealities arising from device updates and peripheral circuits while the modeling of wire parasitics is very simple in Neurosim\cite{neurosim}.  In contrast, \name~considers all circuit and device level non-idealities in more detail. This is especially important when the ratio of synaptic resistance to parasitic resistance is low. TxSim is also capable of evaluating more complex networks compared to CrossSim. When considering simulation time, CrossSim requires about a week to train a simple 3-layer network on MNIST while TxSim is 108x faster for the same task.

\vspace*{-2pt}
\newcolumntype{M}[1]{>{\centering\arraybackslash}m{#1}}
\renewcommand{\arraystretch}{1}
\renewcommand{\tabcolsep}{5pt}
\begin{table}[hbtp]
  \vspace*{-0pt}
  \centering
  \caption{Differentiation with other design tools for training}
  \vspace*{-0pt}
  \begin{tabular}{
  |M{20mm}|M{15mm}|M{18mm}|M{18mm}|}
	\hline
 \textbf{Characteristic} &\textbf{CrossSim} &\textbf{NeuroSim} &\textbf{TxSim}  \\ \hline \hline 
DAC and ADC non-linearity &\cmark &\cmark &\cmark  \\ \hline
Wire parasitics  &\xmark &Simple &Detailed \\ \hline
Update non-linearity   &\cmark &\cmark &\cmark  \\ \hline
Update noise &\cmark &\cmark &\cmark  \\ \hline
Network topology/Dataset   &MLP-MNIST &MLP/CNN/RNN - All datasets &MLP/CNN/RNN - All datasets  \\ \hline
Customizability to other devices and architectures  &Low &Medium &High  \\ \hline

    \end{tabular}    
    
  \vspace*{-10pt}
  \label{tab:lim_control}
\end{table}

\noindent\textbf{Training algorithms.} Given the errors intrinsic in crossbar-based computing, it is important to come up with the right kind of algorithms for training to converge to a good accuracy. To this end, previous efforts~\cite{tayfunCNN,ibmtrain,kendall2020training} propose enhanced algorithms that help overcome non-idealities such as device non-linearity, asymmetry and stochastic noise. Our work complements these efforts by proposing a generic modeling tool that provides an accurate estimate of the degradation due to non-idealities. We expect such tools to further enable future development of crossbar-based training architectures and algorithms.


\begin{figure*}[htb]
  \vspace*{-0pt}
  \centering
  \includegraphics[width=0.8\textwidth]{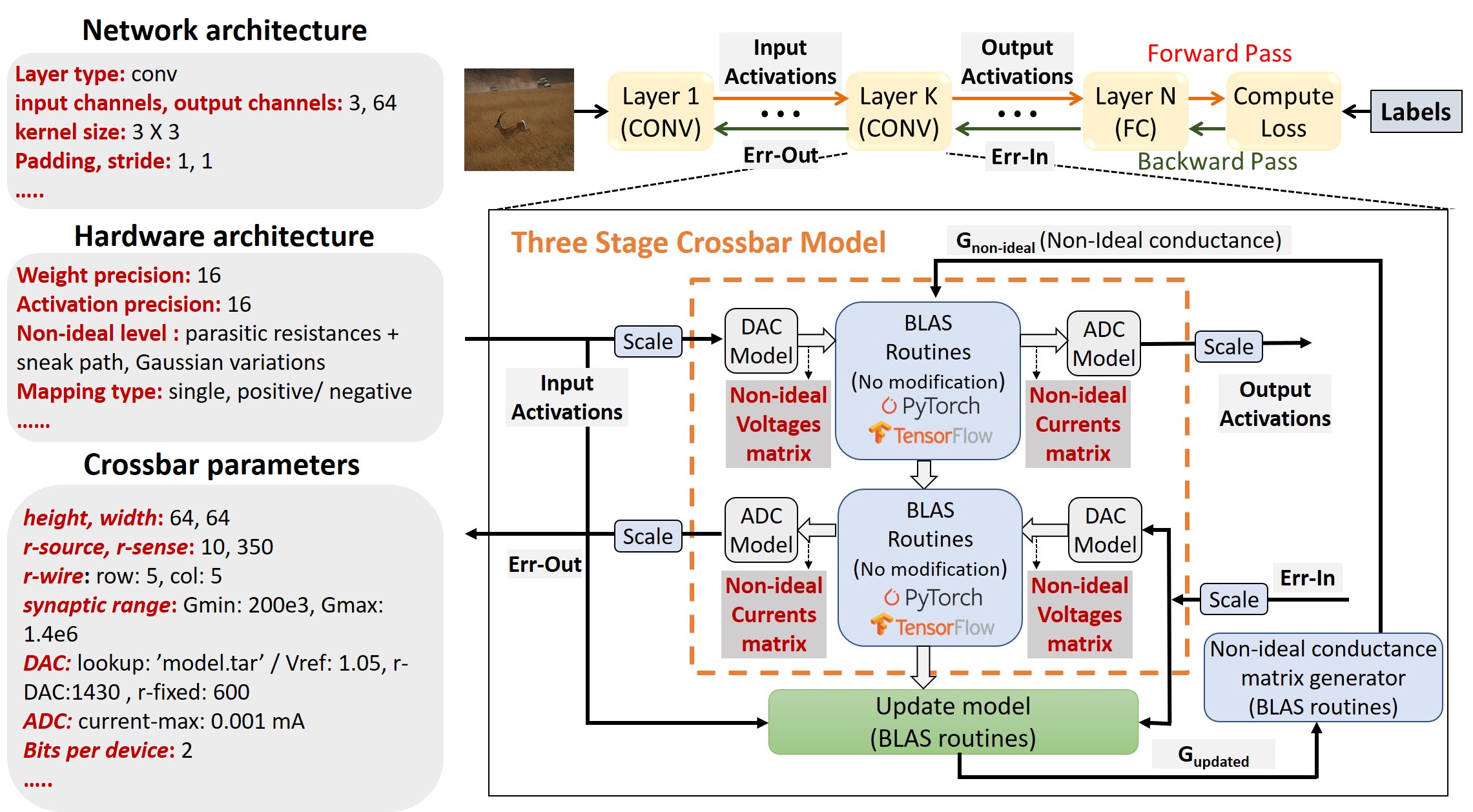}
  \vspace*{-2pt}
  \caption{TxSim framework for evaluating DNN training on crossbar-based systems}
  \label{fig:layout1}
  \vspace*{-8pt}
\end{figure*}


\vspace*{6pt}
\section{Preliminaries}
\label{sec:preliminaries}
\noindent  
In this section, we present a brief overview of DNN training and provide background on resistive crossbars. We then discuss how the non-idealities in the crossbar array impact all phases of training (forward propagation, backward propagation and weight update).

\subsection{DNN Training}
\label{subsec:DNNtrain}

DNN training involves learning weights (strength of the connections) between neurons in each layer in order to match the outputs of the neural network on training samples to the associated labels. The model is typically initialized with random weight parameters that get updated iteratively using Stochastic Gradient Descent (SGD). Different subsets of training data known as minibatches are fed to the model in each iteration and the weights are updated so as to minimize the loss between the model outputs and the labels. The overall training data set is fed to the network multiple times until the loss reaches an optimum value. The performance of the DNN is measured by the total number of correct predictions on unseen test data. DNN training consists of three stages - Forward Propagation, Backward Propagation and Weight Update.

\noindent\textbf{Forward Propagation.} In this stage, the inputs to the DNN are passed through each of its layers in the forward direction to obtain the outputs. For convolutional and fully-connected layers, minibatch inputs are multiplied by the layer weights and passed through non-linear operations such as sigmoid or ReLU to obtain output activations. In addition, other layers such as pooling and batch normalization are also evaluated by applying the appropriate operation on the layer's inputs.\\
\noindent\textbf{Backward Propagation.} The final layer activations determine the loss with respect to the label. The gradient of the loss is computed with respect to weights and activations across the network. The activation gradients are propagated backwards from each layer's output to the layer's input.\\ 
\noindent\textbf{Weight Update.} The gradients of the loss with respect to the network weights are used to compute a weight update. A learning rate is used to control the amount by which the weights get updated.\\

%
\subsection{Resistive Crossbars}
\label{subsec:RCA}
\begin{figure}[htb]
    \centering
    \vspace*{-2pt} 
    \includegraphics[width=0.7\columnwidth]{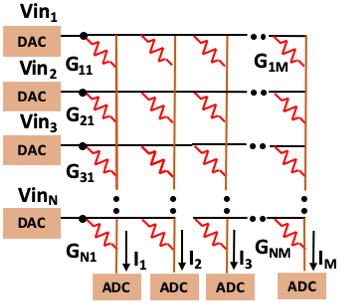}
    \caption{Resistive crossbar array with peripherals}
	\vspace*{-2pt}
    \label{fig:prelim}
\end{figure}
Resistive crossbars are 2D arrays of synaptic elements that can be programmed to store weights and efficiently realize vector matrix multiplications. All the elements in a row of the array are driven by a common input, and all the elements in a column are connected to a common output. The synaptic elements are often realized using non volatile memory (NVM) devices such as PCM and ReRAM. An $N\times M$ resistive crossbar consisting of N rows and M columns is shown in Figure \ref{fig:prelim}. Initially, a write operation is used to program the NVM devices to desired conductance states that represent network weights. The weights of a layer may need to be partitioned and programmed into multiple crossbars. For vector-matrix multiplication, a digital to analog converter (DAC) converts N digital inputs to analog voltages that are fed to the rows. All the rows are activated concurrently and the current from each NVM device is accumulated and sensed at the output of the corresponding column. Finally, the M column currents are converted to digital outputs by passing them through an analog to digital converter (ADC).\\

\subsection{Non-idealities in crossbars}
\label{subsec:RCA}
\noindent\textbf{Peripheral circuitry.} 
The analog computation in the crossbar array requires digital to analog (DAC) and analog to digital
(ADC) converters. The DACs and the ADCs are non-linear and limited in precision to keep their area and power overheads low.\\
\noindent\textbf{Circuit non-idealities.} The wire resistances, source resistances, sink resistances and sneak paths in the crossbar array impact the column currents, causing errors in the vector-matrix multiplication. The errors from the voltage drops across the parasitics and the current from the sneak paths makes the actual current deviate significantly from the ideal output current for a column. \\
\noindent\textbf{Device non-idealities.} The synaptic elements within the crossbar array are inherently stochastic. Existing device technologies can only support limited precisions. These devices also exhibit a non-linear and asymmetric behaviour, suffer from process variations, drift and limited endurance, which can affect the overall classification accuracy.

\vspace*{-0pt}

\vspace*{4pt}
\section{TxSim Modeling framework}
\label{sec:tool}
{\noindent} \name~is a highly customizable and scalable modeling tool that evaluates the application-level accuracy of DNNs trained on crossbar-based hardware. Figure~\ref{fig:layout1} provides an overview of the \name~modeling process. \name~takes three main inputs: (i) the network architecture that defines the number of layers, and the numbers and sizes of input/output channels and kernels, (ii) the hardware architecture parameters such as weight and activation precisions and mapping strategy, and (iii) the crossbar parameters, including DAC/ADC models, crossbar dimensions, synaptic device characteristics, etc. Each layer in the DNN is modeled in \name~as a concatenation of three stages \emph{viz.} the DAC model, the crossbar array model and the ADC model. During the forward pass, the input activations are scaled and sent to the DAC model to obtain the non-ideal voltage matrix. The non-ideal conductance matrix generator converts an ideal conductance matrix to a non-ideal conductance matrix by incorporating all the non-idealities in the core crossbar array. The non-ideal conductance matrix is then multiplied with the non-ideal voltage matrix in the crossbar array model to determine the output currents. Finally, the output currents are passed through the ADC model and scaled to yield output activations. The same procedure is followed in the backward pass as well. During the weight update phase, the input activations and the non-ideal conductance matrix are fed to the update model to obtain G\textsubscript{updated}. 

The rest of the section describes the components of \name~in detail. 

\subsection{Non-ideal conductance matrix generator}
\label{subsec:NIgenerator}

\begin{figure}[htb]
    \centering
    \vspace*{-4pt} 
    \includegraphics[width=\columnwidth]{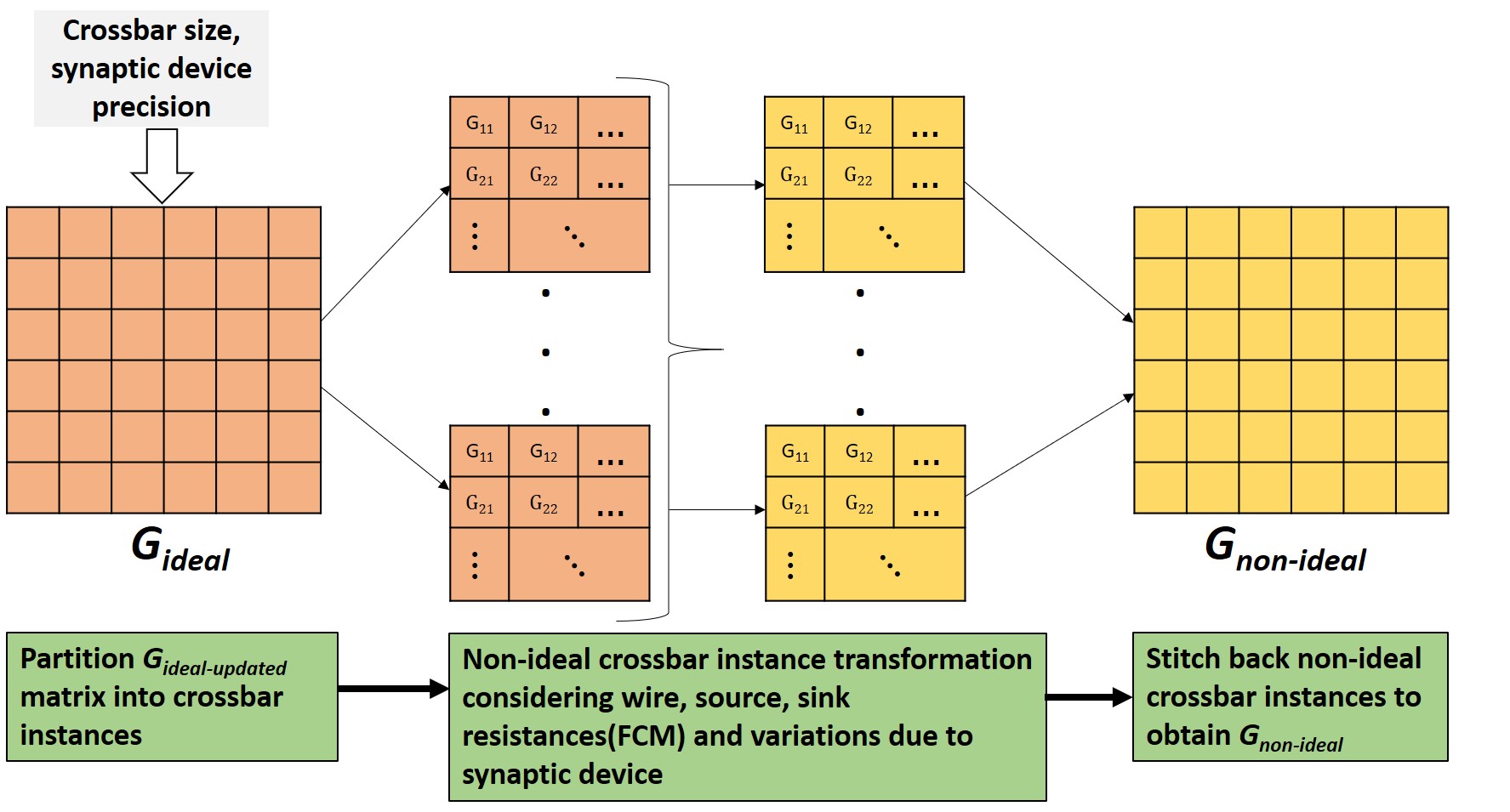}
    \caption{Non-ideal conductance generator}
	\vspace*{-4pt}
    \label{fig:noncondgen}
\end{figure}


The flow of the non-ideal conductance matrix generator is shown in Figure \ref{fig:noncondgen}. The generator analyzes the non-idealities associated with the core crossbar array,~\emph{viz.}, the wire resistances, sink and source resistances, sneak paths, and process variations. It takes an ideal conductance matrix as an input and converts it into a non-ideal conductance matrix that incorporates these core array non-idealities. First, the ideal conductance matrix ($G_{ideal}$) is mapped to one or more synaptic devices based on the on-off ratio and the precision of each device. Next, the ideal conductance matrix is partitioned into crossbar instances based on the specified crossbar dimensions. Within each crossbar instance, the positive and negative conductances may be further mapped onto separate crossbars, obtaining two different currents that are subtracted. Finally, process variations are applied to each synaptic element based on the specified variation profile. The ideal conductance matrices are converted to non-ideal conductance matrices by applying a method called the fast crossbar model (FCM), which was originally proposed for modeling inference~\cite{rxnn}. We provide a brief summary of the method below and refer the reader to ~\cite{rxnn} for a detailed description. 

The FCM conversion mechanism solves a system of equations derived from Ohm's law and Kirchhoff's circuit laws to reflect the impact of non-idealities. An equivalent resistive circuit is first derived for the crossbar array. Each row and column of this resistive circuit is formulated as a row linear system and column linear system and solved independently. The synaptic device is connected to the row and column of the crossbar array at the respective nodes. For each node in the linear system, using Kirchhoff’s Current Law (KCL), the voltage at the node is represented as a function of the voltage across the synaptic device, synaptic conductance and the column or row resistance depending on the linear system. The final non ideal current is the product of the synaptic conductance and the voltage characterized by the linear systems. Key linear algebraic operations such as direct sum, row switching, vector concatenation, row reduction, etc. are used to simplify the matrices.

This process is very accurate and acceptable for inference, since it is performed only once. However, it causes great slowdown when applied to training simulation because it needs to be used after every minibatch iteration to convert the updated weights into non-ideal conductances. Therefore, we propose speedup techniques (discussed in section~\ref{subsec:speedTechniques}) that approximate FCM while maintaining good modeling fidelity. Finally, the crossbar instances are stitched back together to obtain the non-ideal conductance matrix. Note that a copy of the ideal conductance matrix is preserved and used to obtain the $G_{ideal}$ for future minibatch iterations. 


\subsection{Three-stage crossbar model}
\label{subsec:threeStage}
Once we obtain the non-ideal conductance matrix, we utilize a three-stage model (shown in Figure~\ref{fig:layout1}) to perform the forward and backward passes. The incoming digital inputs of each crossbar are converted to voltages depending on the user’s choice of DAC. The voltages and the  non-ideal conductance matrix are fed to the underlying BLAS functions to obtain column currents. Subsequently, the column currents are fed to the ADC model and propagated to the next layer. The maximum current through ADCs is data dependent and obtained by collecting output distribution statistics over multiple training epochs. Peripheral operations such as ReLU, sigmoid, batchnorm, and pooling are computed in the digital domain and are hence unimpaired by crossbar non-idealities. 

\begin{figure}[htb]
    \centering
    \vspace*{-6pt} 
    \includegraphics[width=\columnwidth]{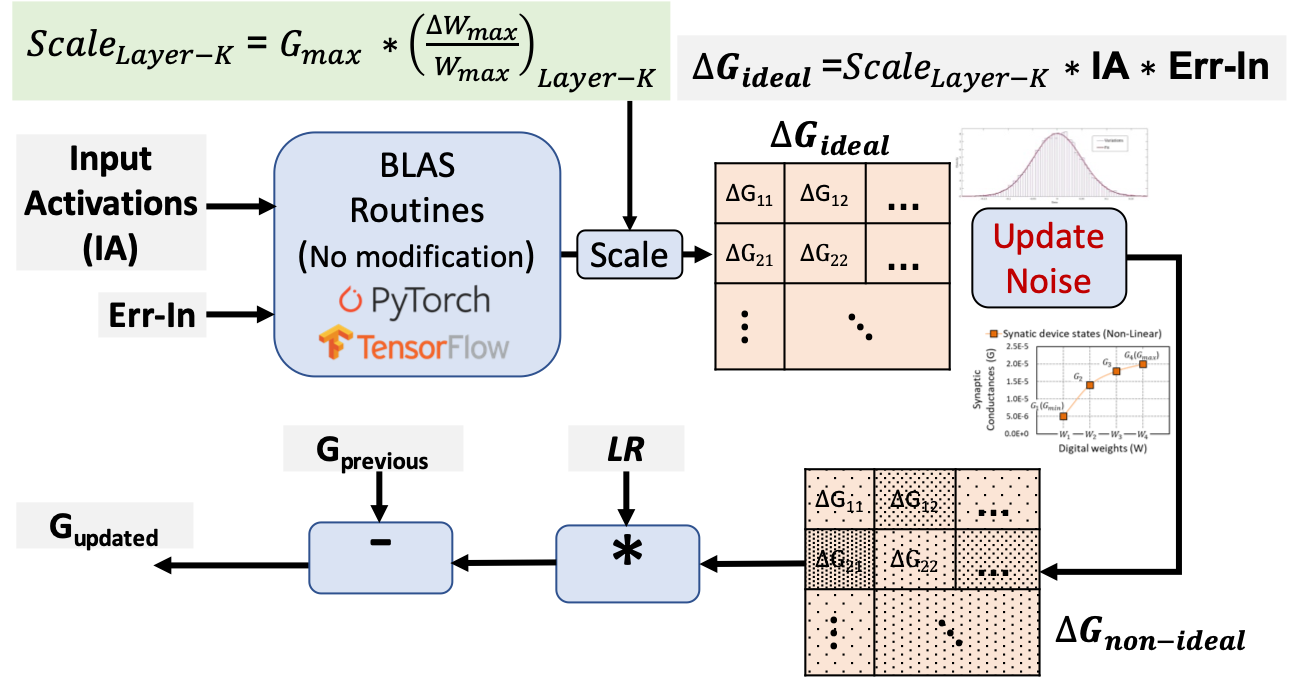}
    \caption{Update noise modeling using TxSim}
	\vspace*{-12pt}
    \label{fig:comp1}
\end{figure}


\subsection{Update model}
\label{subsec:UpdateModel}
The weight update model used \name~is described in Figure~\ref{fig:comp1}.
For efficient weight updates, various parallel update schemes have been proposed~\cite{crossim,ibm_RPU}, wherein inputs and errors are converted to the analog domain and fed to the rows and columns of the crossbar simultaneously. The change in synaptic conductance is proportional to the product of the analog inputs, reflecting the weight update operation. 
The analog inputs may be encoded as time or magnitude based pulses\cite{crossim}, or modeled as stochastic bit streams\cite{ibm_RPU} whose coincidence yields a multiplicative effect. To convert digital gradients to $\Delta$G\textsubscript{ideal}, for every layer, a pre-determined scaling factor ($Scale_{Layer-k}$) is used (shown in Figure~\ref{fig:comp1}). $Scale_{Layer-k}$ is determined using weight and gradient statistics (Wmax and $\Delta$Wmax) collected from native DNN training. The major non-idealities during update operations are the stochastic noise of synaptic devices and the asymmetric write non-linearity~\cite{miti}, which are both modeled in \name.

The update model, shown in equation \eqref{write_NL}, depends on the sign of the update. It depends on the current conductance state (G), the minimum conductance (G\textsubscript{min}), the maximum conductance (G\textsubscript{max}), the ideal conductance change ($\Delta$G\textsubscript{ideal}) and the update non-linearity factor (v). The update non-linearity factor modulates the amount of conductance update written for each positive or negative pulse given the current conductance state. Due to the non-linear nature of the device, the updated conductances deviate from the original values based on v. Another source of non-ideality is the write noise, which arises due to the stochastic nature of the device\cite{crossim}. The conductance change ($\Delta$G\textsubscript{non-ideal}) is sampled from a Gaussian distribution whose standard deviation is $ \gamma*\sqrt{(G\textsubscript{max}-G\textsubscript{min})*G_{ideal}}$, where $\gamma$ is the write noise factor. The write noise is directly proportional to the size of the gradient and a higher write noise factor translates to more write noise being applied to G\textsubscript{ideal}. $\Delta$G\textsubscript{non-ideal} is multiplied by the learning rate (\emph{LR}) and subtracted from the previously stored conductance matrix (G\textsubscript{previous}) in order to obtain the new conductance matrix (G\textsubscript{updated}). G\textsubscript{updated} is passed to the non-ideal conductance matrix generator to obtain the next set of non-ideal conductances for forward propagation. This process is repeated for each DNN layer over multiple epochs until training converges.

\vspace*{-6pt}

\begin{equation} 
\label{write_NL}
\begin{aligned}
\includegraphics[width=0.9\columnwidth]{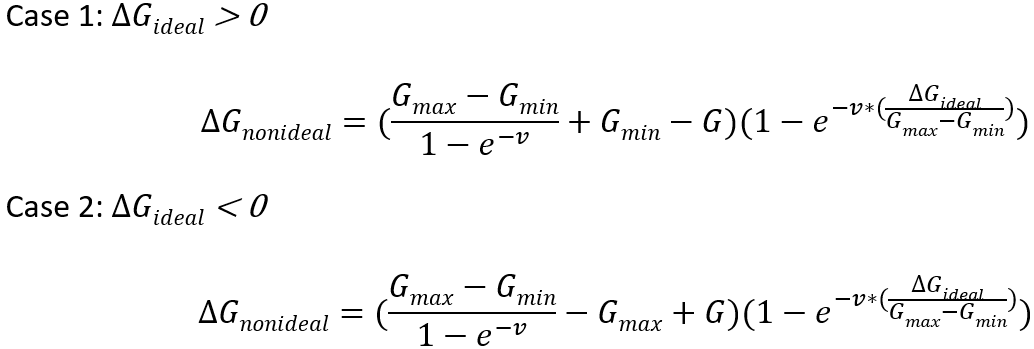}
\end{aligned}
\end{equation}

\subsection{Speedup Techniques}
\label{subsec:speedTechniques}
{\noindent} As mentioned earlier, the generation of the non-ideal conductance matrices is very slow and, while acceptable for inference (where it is one-time), does scale to DNN training (where it needs to be invoked after each minibatch, when weights change). 
Therefore, we present two complementary speedup techniques that significantly accelerate training simulation while preserving good modeling fidelity.

\begin{figure}[htb]
	\vspace*{-10pt}
    \centering
    \includegraphics[width=\columnwidth]{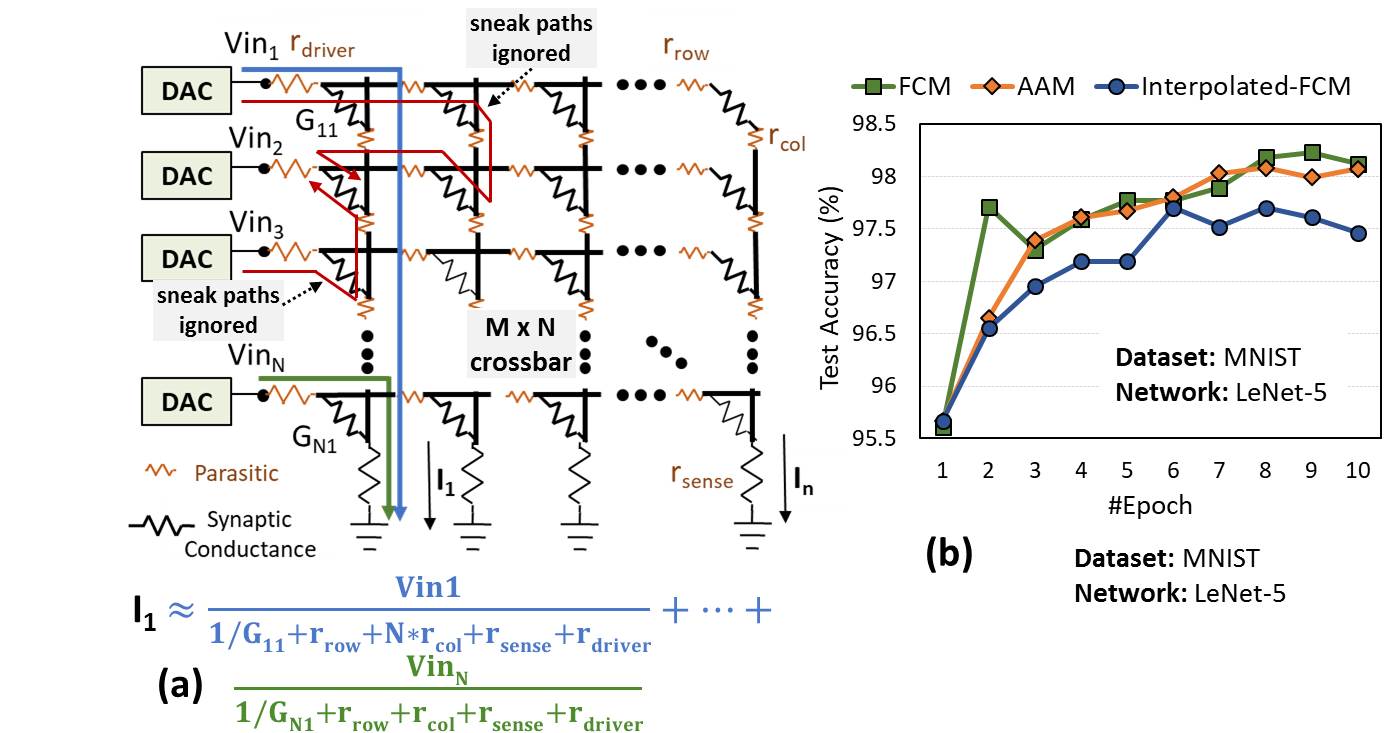}
 	\vspace*{-6pt}
    \caption{(a) Approximate Analytical Model overview, (b) Accuracy comparison for various crossbar array models} 
    \label{fig:approx}
	\vspace*{-6pt}
\end{figure}

\noindent\textbf{Approximate analytical model (AAM).}
\begin{figure}[htb]
	\vspace*{-6pt}
    \centering
    \includegraphics[width=\columnwidth]{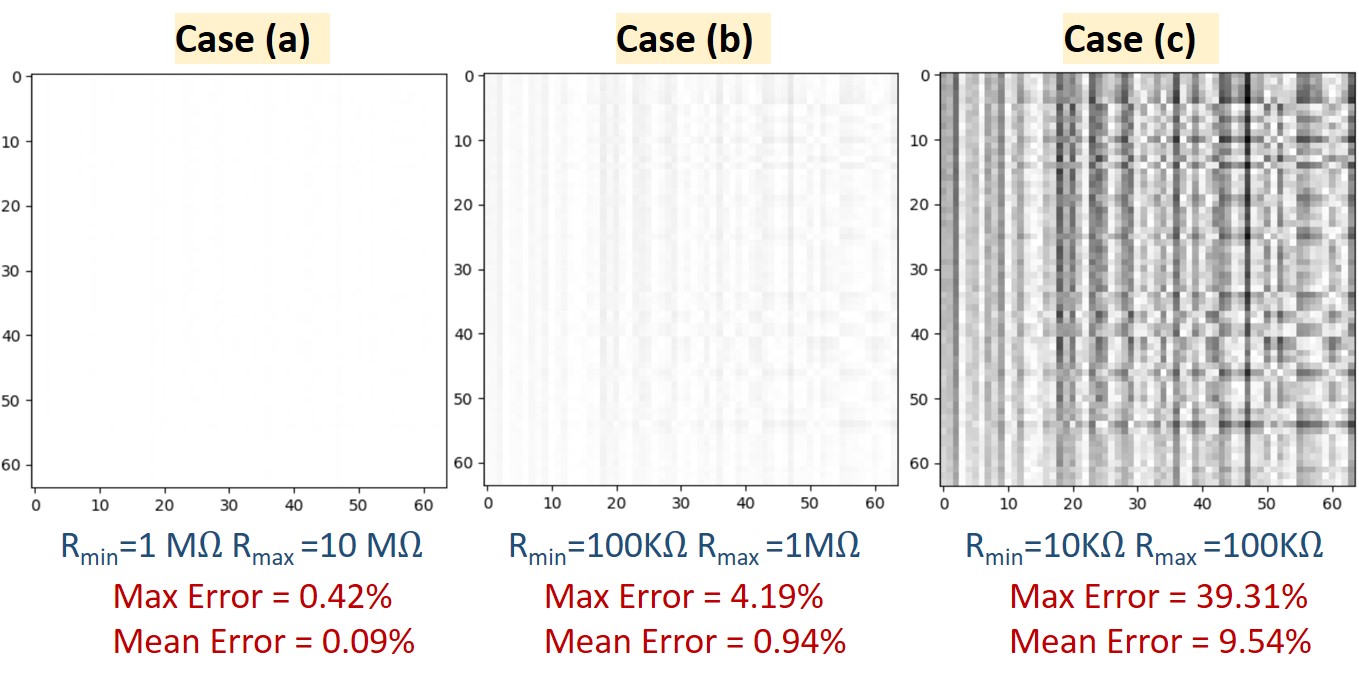}
	\vspace*{-14pt}
    \caption{Error-map of Approximate Analytical model w.r.t FCM for different resistance ranges}
    \label{fig:error}
	\vspace*{-6pt}
\end{figure}
The output current at the column in a non-ideal crossbar can be viewed as a sum of many terms, each corresponding to a path through the crossbar. In the AAM model, we consider a subset of these paths (typically the shorter paths from each row to each column), while ignoring the longer paths since they contribute much smaller currents (as shown in Figure~\ref{fig:approx}(a)). The current for each path is computed considering the source ($r_{source}$), sink ($r_{sense}$), and wire resistances ($r_{row}$ and $r_{col}$). The AAM model allows us to seamlessly trade-off efficiency for accuracy by simply considering more or fewer paths.

We plot the modeling error of AAM with respect to FCM for 64x64 crossbars with different $R_{min}$-$R_{max}$ ranges in Figure~\ref{fig:error}. As shown, AAM is not suitable for case(c) with low $R_{min}$-$R_{max}$ as it results in considerable errors. However, for the higher $R_{min}$-$R_{max}$ range [case (a)], modeling errors are negligible, and in case (b), modeling errors are quite small. Therefore, AAM is used selectively only when the synaptic device resistance range is much higher than wire resistances.

\noindent\textbf{Interpolated-FCM}. In this speedup technique, we perform FCM selectively -- only once every L minibatch iterations (as opposed to each iteration). Every time FCM is performed, the net synaptic conductance distortions due to non-idealities (($G_{ideal}$-$G_{non-ideal}$)/$G_{ideal}$) are computed and stored. For the subsequent L-1 iterations, the $G_{non-ideal}$ is computed using the stored distortion profile. Figure~\ref{fig:approx}(b) shows the application-level accuracy for various models --FCM, AAM, and Interpolated-FCM for the LetNet-5 DNN on the MNIST dataset. As shown, the speedup techniques can effectively model DNN training without much loss in modeling fidelity (note the highly magnified y-axis range).

\vspace*{2pt}
\section{Experimental Methodology}

\label{sec:exptsetup}
{\noindent}

In this section, we briefly describe the methodology used to evaluate \name. The synaptic device used is a Ag/Si ReRAM technology\cite{rrange} with $R_{min}$= 100K$\Omega$, $R_{max}$=1 M$\Omega$. The input voltages to the crossbar are in the range of 0-1V. The DAC and ADC models are calibrated with SPICE based on designs obtained from \cite{verma} and \cite{adc_ibm}. The per-cell row and column resistances are derived from circuit layout and found to be 1\si{\ohm} and 4.6\si{\ohm}, respectively. We conservatively assume 32-bit precision for all data structures (viz.) weights, activations and errors in DNN training based on the scheme proposed in~\cite{suyog}, since it provides classification accuracy close to floating-point training\cite{panther}. Our simulations can be realized using 64x64 crossbar arrays with 2-bit synaptic devices and input streaming through 1-bit DACs.  

{\bf\noindent } 


\vspace*{0pt}

\vspace*{10pt}
\section{Results}
\label{sec:results}
\noindent

In this section, we present results of applying \name~to evaluate the impact of non-idealities on the accuracy of DNNs trained using crossbar-based systems. We also analyze DNN training sensitivity to various device and circuit-level parameters to provide insights for future research.

\subsection{Simulation speed}
\label{subsec:speed}




\begin{wrapfigure}{r}{0.55\columnwidth}
\vspace*{-0pt}
 \begin{center}
  \includegraphics[width=0.55\columnwidth]{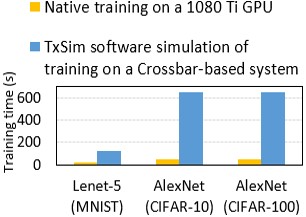}
  \end{center}
  \caption{Training times for Txsim {\em vs.} native training on a GPU}
  \label{fig:slowdown}
  \vspace*{-0pt}
\end{wrapfigure}
To quantify the advantage of \name~in simulation speed, we first compare it to prior frameworks that model training,~\emph{viz.}  CrossSim~\cite{crossim}. We achieve 108x simulation speedup compared to CrossSim~\cite{crossim}. Next, we compare the simulation speed of TxSim to native fixed-point training on a NVIDIA GeForce GTX 1080 Ti GPU. 

For these experiments, we use a batch size of 128 and a crossbar size of 64x64. From Figure \ref{fig:slowdown}, we can observe that software simulation of DNN training on a crossbar-based system is ~14x slower compared to native training on an Nvidia 1080Ti GPU. This is reasonable considering the fact that TxSim emulates crossbar functionality with high modeling fidelity by considering all crossbar non-idealities during the forward, backward, and update operations.


\subsection{Application-level accuracy}
\label{subsec:appAccuracy}

\begin{figure*}[!ht]
  \vspace*{-0pt}
  \centering
  \includegraphics[width=0.9\textwidth]{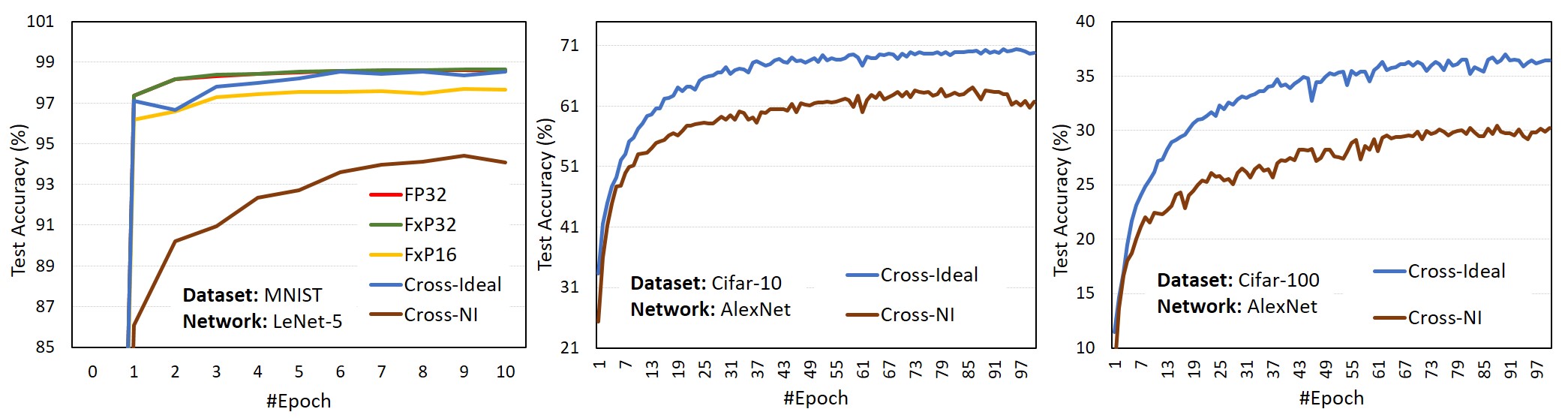}
  \vspace*{-6pt}
  \caption{Impact of crossbar non-idealities on training: LeNet-5 and AlexNet}
  \label{fig:ApplicationResults}
  \vspace*{0pt}
\end{figure*}

\begin{figure}[!h]
    \centering
    \vspace*{-6pt} 
    \includegraphics[width=\columnwidth]{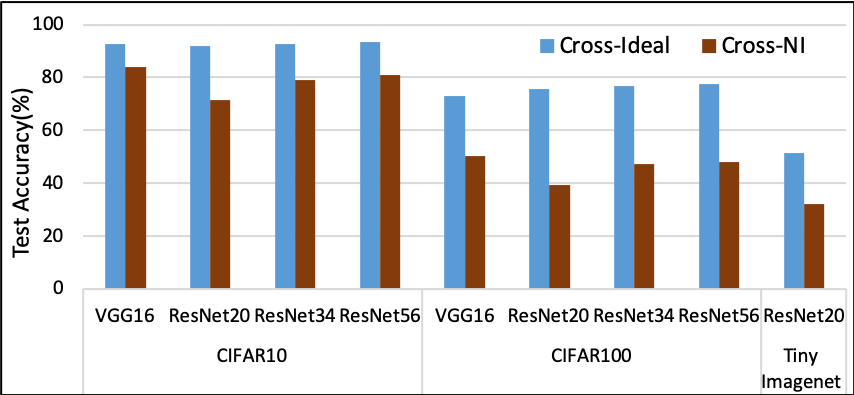}
    \caption{Accuracy degradation due to crossbar non-idealities for larger DNNs}
	\vspace*{-12pt}
    \label{fig:resultslarge}
\end{figure}

\vspace*{-2pt}
\newcolumntype{M}[1]{>{\centering\arraybackslash}m{#1}}
\renewcommand{\arraystretch}{1}
\renewcommand{\tabcolsep}{5pt}



To evaluate the impact of crossbar non-idealities on DNN training, we trained image classification networks on an ideal crossbar system (Cross-Ideal), {\em i.e.}, without any non-idealities, and on a non-ideal crossbar system (Cross-NI) with all crossbar non-idealities. The test accuracy {\em vs.} training epochs for smaller networks such as AlexNet and LeNet-5 is reported in Figure~\ref{fig:ApplicationResults}. As mentioned earlier, We used a 64x64 crossbar with $R_{min}$=100K$\Omega$, $R_{max}$=1M$\Omega$, non-linearity factor (v) = 0.01, and stochastic noise factor ($\gamma$) = 5. In order to have verify the flexibility of \name, we also evaluated a 16-bit Cross-NI system on MNIST. Next, the application level accuracy for larger models such as ResNet-56 on CIFAR100 and ResNet18 on Tiny-ImageNet (subset of ImageNet with 200 classes) is shown in Figure \ref{fig:resultslarge}. The accuracy degradation due to crossbar non-idealities (Cross-NI) is observed to be 3\%-36.4\% across the benchmarks. Moreover, the impact of non-idealities is found to be more prominent on the more complex ResNet-56 and VGG-16 than the simple LeNet-5/AlexNet DNNs. We also observe the accuracy degradation in ResNet-20 to be higher than ResNet-56 and VGG-16. This is because the non-idealities are learnt better while training as the network becomes deeper. Clearly, there is a need to bridge the accuracy gap due to crossbar non-idealities to enable adoption of crossbar-based system for training DNNs. To guide potential solutions to this challenge, we next perform sensitivity analysis to provide insights into the impact of device and circuit-level parameters on accuracy degradation.



\subsection{Sensitivity Analysis}
\label{subsec:SA}
\begin{figure}[H]
  \vspace*{-6pt}
  \centering
  \includegraphics[width=9.1cm,height=4.2cm]{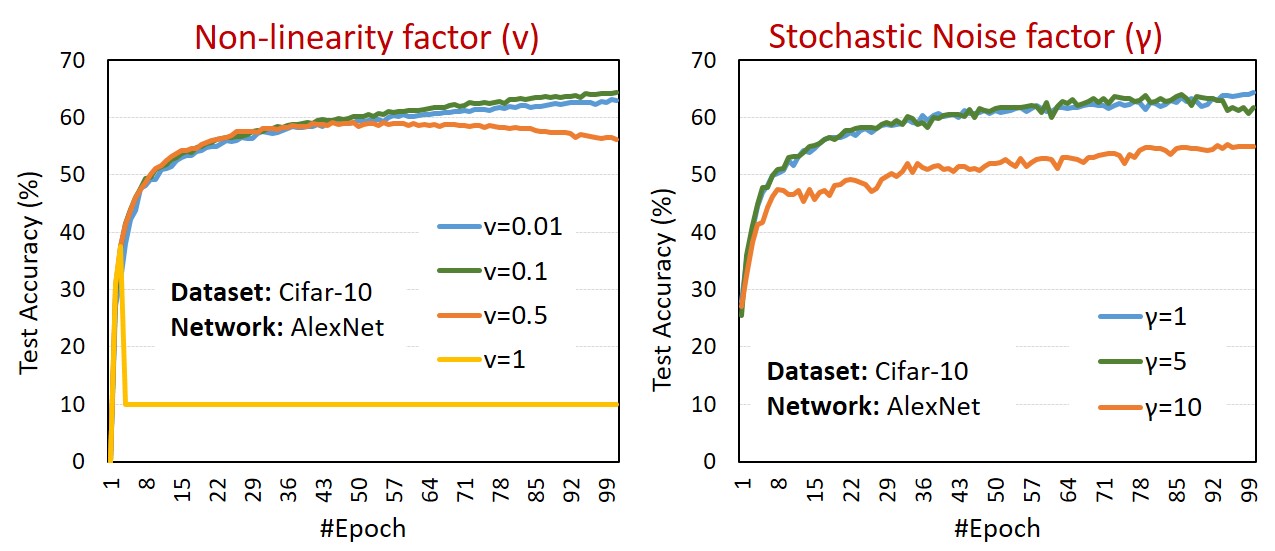}
  \vspace*{-6pt}
  \caption{Sensitivity to update non-idealities}
  \label{fig:sensitivityUpdateNI}
  \vspace*{-8pt}
\end{figure}
\noindent \textbf{Sensitivity to update non-idealities.} Figure~\ref{fig:sensitivityUpdateNI} shows the effect of update non-idealities, viz., write non-linearity and stochastic noise on the application-level accuracy. The crossbar dimensions, on-off ratio and other hardware parameters are kept constant. As non-linearity factor (v) increases from 0.01 to 0.1, there is almost no drop in accuracy. When v is increased to 0.5, the effect of write non-linearity becomes noticeable, and when v is increased to 1, training fails to converge.

Next, the stochastic noise factor ($\gamma$), which determines the standard deviation of the Gaussian distribution from which the write noise is sampled, is varied between 1 to 10. For $\gamma$=1 and $\gamma$=5, the drop in accuracy is almost negligible. However, $\gamma$=10 leads to significant ($\sim$ 10\%) drop in accuracy. 
From these experiments, we conclude that the non-linearity factor should be maintained within 0.1 and the stochastic noise factor within 5 for DNN training on resistive crossbars. 
\begin{floatingfigure}[r]{0.58\columnwidth}
  \vspace*{-0pt} 
  \centering
  \includegraphics[width=0.6\columnwidth]{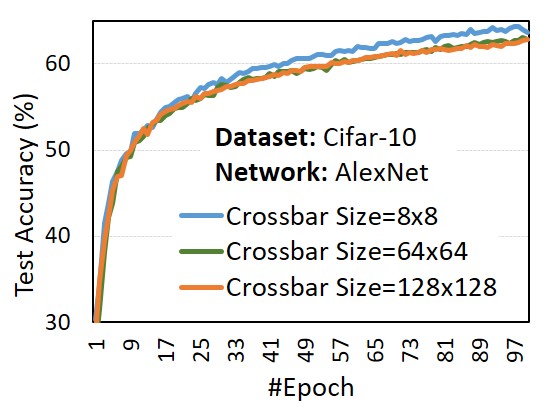}
  \vspace*{-16pt}
  \caption{Sensitivity to crossbar dimensions}
  \label{fig:sensitivityDimen}
  \vspace*{-2pt}
\end{floatingfigure}

\noindent \textbf{Sensitivity to crossbar dimensions.}
Figure~\ref{fig:sensitivityDimen} shows the application-level accuracy with increase in the crossbar dimensions. The accuracy drops slightly for larger crossbars due to increasing impact of all non-idealities including DACs, ADCs, wire resistances and sneak paths. 
In our experiments, the on-off ratio ($R_{max}$/$R_{min}$) of the synaptic device is orders-of-magnitude higher than the wire resistances, highlighting an important observation that the non-idealities due to wire parasitics are less prominent for devices such as ReRAM and PCM. 
However, the effect will be very prominent when the resistance range of the synaptic device is closer to the wire resistances,~\emph{e.g.} Spintronic devices\cite{rxnn,spindle}.

\begin{figure}[H]
  \vspace*{-8pt}
  \centering
  \includegraphics[width=9.1cm,height=4.2cm]{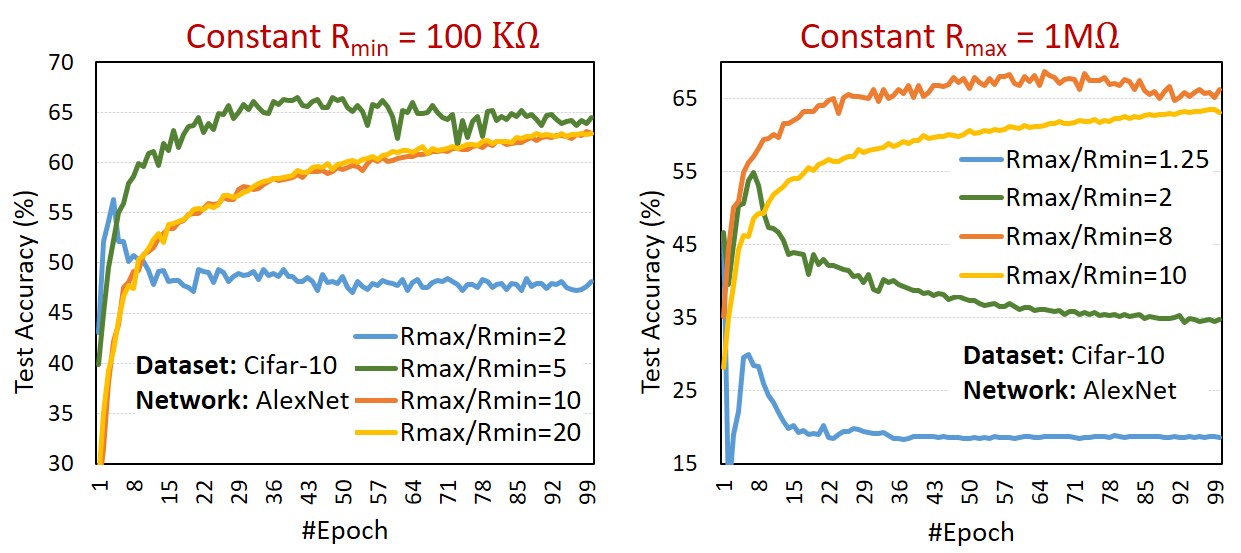}
  \vspace*{-10pt}
  \caption{Sensitivity to $R_{max}$/$R_{min}$ ratio}
  \label{fig:sensitivityON_OFF}
  \vspace*{-6pt}
\end{figure}

\noindent \textbf{Sensitivity to on-off ratio.} To determine the effect of on-off ratio ($R_{max}$/$R_{min}$) on the application-level accuracy, we fix the inputs to the crossbar and change $R_{min}$ and $R_{max}$. We performed two sets of experiments -- (i) increase $R_{max}$ with a fixed $R_{min}$   and (ii) decrease $R_{min}$ with a fixed $R_{max}$.  Increasing the $R_{max}$ values to attain high on-off ratio can make the sensing current low and can lead to a decrease in accuracy due to sensing errors. On the other hand, decreasing $R_{min}$ can have an even greater effect because of circuit non-idealities. Circuit non-idealities have a greater impact when the synaptic device resistance range is close to wire parasitics. For this particular configuration of the device, when R$_{max}$ is 1M$\Omega$, R$_{min}$ can be decreased to maintain a ratio of 8 for the best classification accuracy. Similarly, for the second experiment, the best accuracy is obtained when the on-off ratio is 5. From both the experiments, as indicated in Figure~\ref{fig:sensitivityON_OFF}, we observe that there is a sweet spot in on-off ratio, and ratios that are much smaller or larger can lead to higher accuracy degradation.

In summary, our results indicate that crossbar non-idealities have a major impact impact on the accuracy of DNNs trained on crossbar-based hardware. The accuracy degradation is dependent on various factors such as the crossbar dimensions, non-linearity and stochasticity of the synaptic device, as well its on-off ratio. Future efforts in designing devices,  circuits and architectures for crossbar-based computing must consider these factors. In addition, there is a need for adapting network architectures and training algorithms to minimize the accuracy degradation due to non-idealities.

\vspace*{6pt}
\section{Conclusion}
\label{sec:conclusion}
{\noindent} Crossbar-based systems are extremely promising for efficiently executing DNN training. However, device and circuit-level non-idealities affect application level accuracy significantly. In this work, we propose TxSim, a scalable and customizable modeling tool that evaluates DNN training on resistive crossbars considering the impact of all computational non-idealities. TxSim models a more comprehensive set of non-idealities than prior works while also achieving 6x-108x speedup over prior frameworks. To further improve the simulation runtime for complex datasets and network architectures, we also propose speedup techniques, viz., approximate analytical model (AAM) and interpolated FCM, that show a good balance between modeling fidelity and simulation runtime. Using TxSim, we evaluate several DNN benchmarks and observe that the accuracy degradation can be considerable (3\%-36.4\%). We also perform sensitivity analysis to gain further insights into the impact of various circuit and device level parameters on crossbar-based DNN training.

\section{Acknowledgment}
This work was supported by C-BRIC, one of six centers in JUMP, a Semiconductor Research Corporation (SRC) program, sponsored by DARPA.
\vspace*{-0pt}
\scriptsize
\bibliographystyle{unsrt}
\bibliography{paper}

\begin{thebibliography}{10}

\bibitem{fortune-dnns}
R.~Parloff.
\newblock {The AI Revolution: Why Deep Learning Is Suddenly Changing Your Life.
  http://fortune.com/ai-artificial-intelligence-deep-machine-learning/ }.
\newblock Online. Accessed Sept. 17, 2017.

\bibitem{speech}
Awni Hannun, Carl Case, Jared Casper, Bryan Catanzaro, Greg Diamos, Erich
  Elsen, Ryan Prenger, Sanjeev Satheesh, Shubho Sengupta, Adam Coates, and
  Andrew~Y. Ng.
\newblock Deep speech: Scaling up end-to-end speech recognition, 2014.

\bibitem{tpu}
N.P.~Jouppi et~al.
\newblock In-datacenter performance analysis of a tensor processing unit.
\newblock In {\em Proceedings of the 44th Annual International Symposium on
  Computer Architecture}, ISCA ’17, page 1–12, New York, NY, USA, 2017.
  Association for Computing Machinery.

\bibitem{brainwave}
Jeremy Fowers, Kalin Ovtcharov, Michael Papamichael, Todd Massengill, Ming Liu,
  Daniel Lo, Shlomi Alkalay, Michael Haselman, Logan Adams, Mahdi Ghandi,
  Stephen Heil, Prerak Patel, Adam Sapek, Gabriel Weisz, Lisa Woods, Sitaram
  Lanka, Steven~K. Reinhardt, Adrian~M. Caulfield, Eric~S. Chung, and Doug
  Burger.
\newblock A configurable cloud-scale dnn processor for real-time ai.
\newblock In {\em Proceedings of the 45th Annual International Symposium on
  Computer Architecture}, ISCA ’18, page 1–14. IEEE Press, 2018.

\bibitem{scaledeep}
Swagath Venkataramani, Ashish Ranjan, Subarno Banerjee, Dipankar Das, Sasikanth
  Avancha, Ashok Jagannathan, Ajaya Durg, Dheemanth Nagaraj, Bharat Kaul,
  Pradeep Dubey, and Anand Raghunathan.
\newblock Scaledeep: A scalable compute architecture for learning and
  evaluating deep networks.
\newblock {\em SIGARCH Comput. Archit. News}, 45(2):13–26, June 2017.

\bibitem{nas}
Thomas Elsken, Jan~Hendrik Metzen, and Frank Hutter.
\newblock Neural architecture search: A survey.
\newblock {\em Journal of Machine Learning Research}, 20(55):1--21, 2019.

\bibitem{pcm}
B.~{Rajendran}, H.~{Lung}, and C.~{Lam}.
\newblock Phase change memory — opportunities and challenges.
\newblock In {\em 2007 International Workshop on Physics of Semiconductor
  Devices}, pages 92--95, 2007.

\bibitem{reram}
H.Akinaga et~al.
\newblock {Resistive Random Access Memory ({R}e{RAM}) Based on Metal Oxides}.
\newblock 2010.

\bibitem{neuromorphicHardware_Survey}
Catherine~D. Schuman, Thomas~E. Potok, Robert~M. Patton, J.~Douglas Birdwell,
  Mark~E. Dean, Garrett~S. Rose, and James~S. Plank.
\newblock A survey of neuromorphic computing and neural networks in hardware.
\newblock {\em CoRR}, abs/1705.06963, 2017.

\bibitem{ibmall}
Shubham Jain, Aayush Ankit, Indranil Chakraborty, Tayfun Gokmen, Malte~J.
  Rasch, Wilfried Haensch, Kairshik Roy, and Anand Raghunathan.
\newblock Neural network accelerator design with resistive crossbars:
  Opportunities and challenges.
\newblock {\em IBM J. Res. Dev.}, 63:10:1--10:13, 2019.

\bibitem{pipelayer}
Linghao Song, Xuehai Qian, Hai Li, and Yiran Chen.
\newblock Pipelayer: A pipelined reram-based accelerator for deep learning.
\newblock {\em 2017 IEEE International Symposium on High Performance Computer
  Architecture (HPCA)}, pages 541--552, 2017.

\bibitem{crossim}
Sapan Agarwal, Steven~J. Plimpton, David~R. Hughart, Alexander~H. Hsia, Isaac
  Richter, Jonathan~A. Cox, Conrad~D. James, and Matthew~J. Marinella.
\newblock Resistive memory device requirements for a neural algorithm
  accelerator.
\newblock {\em 2016 International Joint Conference on Neural Networks (IJCNN)},
  pages 929--938, 2016.

\bibitem{tayfunMLP}
G{\"o}kmen Tayfun and Yurii Vlasov.
\newblock Acceleration of deep neural network training with resistive
  cross-point devices: Design considerations.
\newblock {\em Frontiers in Neuroscience}, 10, 2016.

\bibitem{neurosim}
Pai-Yu Chen, Xiaochen Peng, and Shimeng Yu.
\newblock Neurosim: A circuit-level macro model for benchmarking neuro-inspired
  architectures in online learning.
\newblock {\em IEEE Transactions on Computer-Aided Design of Integrated
  Circuits and Systems}, 37:3067--3080, 2018.

\bibitem{pytorx}
Zhezhi et~al.
\newblock Noise injection adaption: End-to-end reram crossbar non-ideal effect
  adaption for neural network mapping.
\newblock In {\em Proc DAC 2019}.

\bibitem{rxnn}
S.~{Jain}, A.~{Sengupta}, K.~{Roy}, and A.~{Raghunathan}.
\newblock {RxNN: A Framework for Evaluating Deep Neural Networks on Resistive
  Crossbars}.
\newblock {\em IEEE Transactions on Computer-Aided Design of Integrated
  Circuits and Systems}, pages 1--1, 2020.

\bibitem{mnsim2}
Zhenhua Zhu, Hanbo Sun, Kaizhong Qiu, Lixue Xia, Gokul Krishnan, Guohao Dai,
  Dimin Niu, Xiaoming Chen, X.~Sharon Hu, Yu~Cao, Yuan Xie, Yu~Wang, and
  Huazhong Yang.
\newblock {\em MNSIM 2.0: A Behavior-Level Modeling Tool for Memristor-Based
  Neuromorphic Computing Systems}, page 83–88.
\newblock Association for Computing Machinery, New York, NY, USA, 2020.

\bibitem{noiseinj}
Zhezhi He, Jie Lin, Rickard Ewetz, Jiann-Shiun Yuan, and Deliang Fan.
\newblock Noise injection adaption: End-to-end reram crossbar non-ideal effect
  adaption for neural network mapping.
\newblock In {\em Proceedings of the 56th Annual Design Automation Conference
  2019}, DAC '19, New York, NY, USA, 2019. Association for Computing Machinery.

\bibitem{Chakraborty_2018}
Indranil Chakraborty, Deboleena Roy, and Kaushik Roy.
\newblock Technology aware training in memristive neuromorphic systems for
  nonideal synaptic crossbars.
\newblock {\em IEEE Transactions on Emerging Topics in Computational
  Intelligence}, 2:335--344, 2018.

\bibitem{cdnn}
Shubham Jain and Anand Raghunathan.
\newblock Cxdnn: Hardware-software compensation methods for deep neural
  networks on resistive crossbar systems.
\newblock {\em ACM Trans. Embed. Comput. Syst.}, 18(6), November 2019.

\bibitem{mnsim}
L.~{Xia}, B.~{Li}, T.~{Tang}, P.~{Gu}, P.~{Chen}, S.~{Yu}, Y.~{Cao}, Y.~{Wang},
  Y.~{Xie}, and H.~{Yang}.
\newblock Mnsim: Simulation platform for memristor-based neuromorphic computing
  system.
\newblock {\em IEEE Transactions on Computer-Aided Design of Integrated
  Circuits and Systems}, 37(5):1009--1022, 2018.

\bibitem{tayfunCNN}
{T. Gokmen et al.}
\newblock Training deep convolutional neural networks with resistive
  cross-point devices.
\newblock {\em Frontiers in Neuroscience}, 2017.

\bibitem{ibmtrain}
M.J.~Rasch et~al.
\newblock Training large-scale {ANN}s on simulated resistive crossbar arrays,
  2019.

\bibitem{kendall2020training}
Jack Kendall, Ross Pantone, Kalpana Manickavasagam, Yoshua Bengio, and Benjamin
  Scellier.
\newblock Training end-to-end analog neural networks with equilibrium
  propagation, 2020.

\bibitem{ibm_RPU}
Seyoung Kim, Tayfun Gokmen, Hyung-Min Lee, and Wilfried~E. Haensch.
\newblock Analog cmos-based resistive processing unit for deep neural network
  training.
\newblock {\em 2017 IEEE 60th International Midwest Symposium on Circuits and
  Systems (MWSCAS)}, pages 422--425, 2017.

\bibitem{miti}
Pai-Yu Chen, Binbin Lin, I-Ting Wang, Tuo-Hung Hou, Jieping Ye, Sarma Vrudhula,
  Jae-sun Seo, Yu~Cao, and Shimeng Yu.
\newblock Mitigating effects of non-ideal synaptic device characteristics for
  on-chip learning.
\newblock In {\em Proceedings of the IEEE/ACM International Conference on
  Computer-Aided Design}, ICCAD '15, page 194–199. IEEE Press, 2015.

\bibitem{rrange}
Kuk-Hwan Kim, Siddharth Gaba, Dana~C. Wheeler, Jose~M. Cruz-Albrecht, Tahir
  Hussain, Narayan Srinivasa, and Wei Lu.
\newblock A functional hybrid memristor crossbar-array/cmos system for data
  storage and neuromorphic applications.
\newblock {\em Nano letters}, 12 1:389--95, 2012.

\bibitem{verma}
{Jintao Zhang}, {Zhuo Wang}, and N.~{Verma}.
\newblock A machine-learning classifier implemented in a standard 6t sram
  array.
\newblock In {\em 2016 IEEE Symposium on VLSI Circuits (VLSI-Circuits)}, pages
  1--2, 2016.

\bibitem{adc_ibm}
Jing Li, Chao-I Wu, Scott~C. Lewis, Jackie Morrish, Tien-Yen Wang, Richard
  Jordan, Tom Maffitt, Matthew~J. Breitwisch, Alejandro~G. Schrott, Roger
  Cheek, Hsiang-Lan Lung, and Chung Lam.
\newblock A novel reconfigurable sensing scheme for variable level storage in
  phase change memory.
\newblock {\em 2011 3rd IEEE International Memory Workshop (IMW)}, pages 1--4,
  2011.

\bibitem{suyog}
Suyog Gupta, Ankur Agrawal, Kailash Gopalakrishnan, and Pritish Narayanan.
\newblock Deep learning with limited numerical precision.
\newblock In {\em Proceedings of the 32nd International Conference on
  International Conference on Machine Learning - Volume 37}, ICML’15, page
  1737–1746. JMLR.org, 2015.

\bibitem{panther}
Aayush Ankit, Izzat~El Hajj, Sai~Rahul Chalamalasetti, Sapan Agarwal, Matthew
  Marinella, Martin Foltin, John~Paul Strachan, Dejan Milojicic, Wen mei Hwu,
  and Kaushik Roy.
\newblock Panther: A programmable architecture for neural network training
  harnessing energy-efficient reram, 2019.

\bibitem{spindle}
Shankar~Ganesh Ramasubramanian, Rangharajan Venkatesan, Mrigank Sharad, Kaushik
  Roy, and Anand Raghunathan.
\newblock Spindle: Spintronic deep learning engine for large-scale neuromorphic
  computing.
\newblock In {\em Proceedings of the 2014 International Symposium on Low Power
  Electronics and Design}, ISLPED '14, page 15–20, New York, NY, USA, 2014.
  Association for Computing Machinery.

\end{thebibliography}

\end{document}